%% file: main.tex
\documentclass[11pt]{article}

\usepackage[a4paper,margin=1in]{geometry}
\usepackage[T1]{fontenc}
\usepackage[utf8]{inputenc}
\usepackage{lmodern}
\usepackage[expansion=false]{microtype} 
\usepackage{amsmath,amssymb}
\usepackage{graphicx}
\usepackage{booktabs}
\usepackage{pdflscape} 
\usepackage{array}
\newcolumntype{L}[1]{>{\raggedright\arraybackslash}p{#1}}
\usepackage{caption}
\usepackage{enumitem}
\usepackage[section,above]{placeins} 
\usepackage{xcolor}
\usepackage[skins,breakable]{tcolorbox}
\usepackage[numbers,sort&compress]{natbib}
\usepackage[hidelinks]{hyperref}
\hypersetup{pdftitle={Harness Tokenomics: A Router for the Enterprise Agentic Control Plane},
  pdfauthor={Ted Kwartler, Alan Aqrawi, Arian Abbasi}}
\usepackage{url}

\graphicspath{{figures/}}
\setlist{nosep,leftmargin=*,after={\suppressfloats[t]}} 
\definecolor{accent}{HTML}{1F3A93}
\definecolor{boxbg}{HTML}{F6F6F3}
\newtcolorbox{summarybox}{enhanced,breakable,before skip=6pt,colback=boxbg,colframe=accent,boxrule=0.6pt,arc=2pt,
  left=7pt,right=7pt,top=5pt,bottom=5pt,fonttitle=\bfseries,fontupper=\small,title=Key findings,
  before upper={\setlength{\parskip}{2.5pt}}}
\newcommand{\keynum}[2]{\begin{minipage}[t]{0.315\linewidth}\centering
  {\color{accent}\bfseries\fontsize{19}{21}\selectfont #1}\\[2pt]{\footnotesize #2}\end{minipage}}

\newcommand{\Rh}{R_{\mathrm{h}}}
\newcommand{\Rt}{R_{\mathrm{t}}}
\newcommand{\Wt}{W_{\mathrm{t}}}
\newcommand{\Wh}{W_{\mathrm{h}}}

\title{\vspace{-3.2em}\textbf{Harness Tokenomics}\\[4pt]
\large A Router for the Enterprise Agentic Control Plane}
\author{Ted Kwartler\textsuperscript{1,2,*}\qquad Alan Aqrawi\textsuperscript{1}\qquad Arian Abbasi\textsuperscript{1}\\[3pt]
\footnotesize\textsuperscript{1}Accenture\qquad \textsuperscript{2}Harvard University, Faculty of Arts and Sciences (FAS)\\
\footnotesize\textsuperscript{*}\href{mailto:edwardkwartler@fas.harvard.edu}{edwardkwartler@fas.harvard.edu},
\href{mailto:ted.kwartler@accenture.com}{ted.kwartler@accenture.com}}
\date{\vspace{-2.0em}}

\begin{document}
\maketitle
\kern-1.2em 

\begin{abstract}
\input{sections/00_abstract}
\end{abstract}

\input{sections/00_summary}
\input{sections/01_intro}
\input{sections/02_background}
\input{sections/03_problem}
\input{sections/04_routing}
\input{sections/05_casestudy}
\input{sections/06_actions}
\input{sections/07_related}
\input{sections/08_limitations}
\input{sections/09_conclusion}

\paragraph{Reproducibility.} Appendix~\ref{app:taxonomy} specifies the taxonomy and policy and
Appendix~\ref{app:method} the cost model and its constants, so that any enterprise can rebuild the method
with its own inputs. The authors' simulator, scenario catalogue and cached classifier labels are not
released.

\paragraph{Disclaimer.} The spend, prices, populations and savings in this paper come from public list prices,
public session corpora, the authors' own sessions and an emulated enterprise. They do not represent Accenture's spend, prices, contract
terms or internal figures.

\bibliographystyle{unsrtnat}
\bibliography{refs}

\appendix
\input{sections/A_taxonomy}
\input{sections/B_method}
\input{sections/C_landscape}
\input{sections/D_figures}

\end{document}

%% file: sections/00_abstract.tex
Harnesses, the products that run AI coding agents, are multiplying, and enterprises are rolling them out to their employees. What started as pilots with a few hundred seats is now scaling to tens of thousands. Enterprises rarely build these harnesses and usually adopt the ones model vendors bundle with their seats and APIs, such as Anthropic's Claude Code or OpenAI's Codex. A harness decides which model answers, what the model reads, how the prompt cache is used and which subagents run, so it picks the rate on the price sheet and sets the volume bought at it. Enterprises that keep a proprietary or untuned harness at its defaults inherit these choices and their bill. We build a fast, customisable router in which Jev, a classifier with calibrated probabilities, labels every prompt against a bring-your-own taxonomy of agentic requests. One user turn is many requests over a prompt cache that belongs to one model, so the router moves work only where no running conversation has to rebuild its cache. It routes at session start, in side lanes and at subagent launch. From the price sheet we derive when a mid-task switch pays back. We also find a crossover, in which the highest-priced model costs less than the next tier on long tool-heavy sessions, and repricing about 10,000 real sessions from public datasets confirms it. In an emulated enterprise of 10,000 seats with user behaviour taken from these datasets, the router recovers 13 to 21\% of model spend at Anthropic's list prices of 21 September 2026, \$3.3M to \$5.1M a year. The paper also maps the risks across twenty-one harnesses, prices the dependence on one vendor's models, and proposes a control plane that enterprises can run from within, starting now, with a ladder for deciding later whether to own the harness.

%% file: sections/00_summary.tex
\begin{summarybox}
\noindent\keynum{13--21\%}{of agentic model spend recovered by cache-safe routing}\hfill
\keynum{\$3.3--5.1M}{saved a year by a 10,000-seat enterprise}\hfill
\keynum{51--80\%}{of long real sessions cost less on top-tier Fable 5.1 than on Opus 5}

\smallskip
\textbf{The harness sets the bill, and routing cuts it by 13 to 21\%.} We emulated an enterprise of 10,000 seats, with corrections, pauses and subagents as measured in public datasets. It spends \$24.4M a year on models, \$41 per engineer per active day. Sending each task to the cheapest capable model, without disturbing work in progress, saves \$3.3M a year with today's classifier and up to \$5.1M with a perfect one. A one-line change in a long coding session drops from \$0.90 to \$0.36.

\textbf{List price is the wrong metric.} On long tool-heavy sessions the top model re-reads the conversation at half the price of the next tier and costs less in 51 to 80\% of real sessions over \$100. A mid-task switch that later returns rebuilds the cache twice and needs one to eleven turns to pay off.

\textbf{Pauses matter as much as the classifier.} The API cache lasts five minutes, and 17 to 27\% of user turns follow a longer pause and pay for the whole conversation again. The saving runs from 23\% without such pauses to 8\% for users who pause most, and a one-hour cache adds two points.

\textbf{Move work only when sure.} Corrections, prompts that fix an earlier answer, are over a third of that \$24.4M, and a task given to too weak a model triggers more. The router acts only on confident labels and never downgrades a correction.

\textbf{Take control from within, starting this quarter.} Put a cache-safe gateway in front of every harness to pick the model for each session, side task and subagent, never above the user's own pick. Turn on the one-hour cache, report token use only in aggregate, let each function define its own task categories and ship one short bundle of procedures and connectors. Decide later, on purpose, whether to own the harness. Harnesses that accept the enterprise's own endpoint also reach other vendors' strong models, at 0.12 to 0.80 times the cost of Opus 5.
\end{summarybox}

%% file: sections/01_intro.tex
\section{Introduction}\label{sec:intro}

Agents are becoming the way enterprises use frontier models. A harness sends the model a task, runs the tools it asks for, feeds the results back and repeats until the work is done. Claude Code, Codex, Gemini CLI, OpenCode and more than a dozen others ship this loop as a product, and enterprises are moving them from pilots to production. Accenture announced training on Claude for about 30,000 professionals and Claude Code for tens of thousands of its developers~\citep{accenture_anthropic2025}, Cognizant Claude for up to 350,000 employees~\citep{cognizant_anthropic2025}, and PwC a rollout of Claude Code and Cowork that starts with its U.S. teams and expands toward a global workforce of hundreds of thousands~\citep{pwc_anthropic2026}. The work grows with them: GitHub's Copilot coding agent alone opened more than a million pull requests between May and September 2025~\citep{octoverse2025}, and the share of Codex users who sent at least one prompt estimated to need an hour or more of skilled work doubled in five months, from 35 to 70\%~\citep{codexdiff2026}. In 2025 most developers had yet to adopt agents~\citep{so2025}, so the defaults an enterprise accepts now will govern most of its agentic work to come.

Enterprises rarely build these harnesses and usually adopt the ones model vendors bundle with their seats and APIs. The harness has no licence fee of its own. It comes with a Claude or ChatGPT seat or runs on an API key, and the tokens it spends are billed, on Anthropic's Enterprise plan every one at API rates~\citep{claude_enterprise_billing,codex_pricing}. With the harness come its defaults, and Anthropic's own usage data suggests they weigh heavily: on the same model, Claude Code sessions run more autonomously than chat, ``suggesting that the product used is likely more important than the underlying model''~\citep{aei_jun26}. The harness chooses the model for every request, writes the prompts, manages the prompt cache, launches subagents and decides what is logged. The price sheet sets what a token costs, and the harness decides which rate applies and how many tokens are bought at it, so it shapes both what the work costs and who controls it. In Claude Code the defaults are the vendor's own models, subagents on the parent's model and a five-minute cache for API customers. Section~\ref{sec:problem} names the four risks this hands to the enterprise: coupling to one vendor's prices, closed code, skills and connectors fragmented across harnesses, and dependence on individual user skill.

Cost is where the defaults show first, and routing is the obvious lever: send each task to the cheapest model that can do it. Chat routers decide per request, and that fails inside a harness. There one user turn is many requests over a prompt cache that belongs to one model, and moving a turn to another model pays to rebuild the cache. We make routing work inside the harness. From the price sheet we derive after how many requests a mid-task switch pays back (Section~\ref{sec:payback}). Under Anthropic's list prices of 21 September 2026 the highest-priced model also costs less per session than the next tier for long tool-heavy work, a crossover that repricing about 10,000 real sessions confirms (Section~\ref{sec:crossover}). The router therefore moves work only where no running conversation has to rebuild its cache: it chooses the model at session start, runs small tasks in separate lanes and routes subagents at launch. Jev, a classifier built for fast structured decisions, labels every prompt against a bring-your-own taxonomy of agentic requests in under half a second, and the router moves a turn only when Jev's calibrated probabilities say the cheaper model suffices (Section~\ref{sec:routing}).

In a case study that emulates a 10,000-seat enterprise with user behaviour measured in real sessions, the router recovers 13 to 21\% of model spend, \$3.3M to \$5.1M a year, and users' pauses move the saving as much as the classifier does (Section~\ref{sec:results}). The cost model takes any seat count. At 30,000 seats the saving is \$9.8M to \$15.2M a year, and at 100,000 seats \$32.6M to \$50.6M. The harness also decides which vendors are in reach. On the same real sessions, strong models of other vendors cost 0.12 to 0.80 times Opus 5, and Claude Code, which works with Anthropic's models only, cannot call them. Section~\ref{sec:actions} turns these results into a control plane that an enterprise can run from within in the short term, and a ladder for deciding later, on purpose, whether to own the harness.

\paragraph{Contributions.} The paper contributes:
\begin{enumerate}
\item A cost model of agentic turns under model-scoped caching: the payback rule for mid-task switches and the crossover result, checked against real sessions.
\item A cache-safe router: execution modes that route at session start, in side lanes and at subagent launch, and a policy that acts on a fast classifier's calibrated probabilities over a bring-your-own taxonomy.
\item The case study with its sensitivities, the price of lock-in on real sessions, a control-plane architecture and a sovereignty ladder.
\item An emulation method for cost cases before telemetry exists: a tagged catalogue perturbed with behaviour measured in real sessions, a live classifier, misroutes charged as correction chains, population scaling.
\item A risk map of third-party agentic harnesses, with a coupling matrix over twenty-one harnesses built from licence files and configuration documents (Table~\ref{tab:landscape}).
\end{enumerate}
Section~\ref{sec:background} explains how harnesses work and who ships them, Section~\ref{sec:problem} states the problem, Section~\ref{sec:routing} derives the cost rules and specifies the router and its emulation, Section~\ref{sec:results} reports the results, Section~\ref{sec:actions} turns them into recommendations, Section~\ref{sec:related} places the work and Section~\ref{sec:limits} states its limits.

%% file: sections/02_background.tex
\section{Background}\label{sec:background}

\subsection{Every tool step re-sends the conversation}\label{sec:mechanics}

\paragraph{The loop.} A harness holds a conversation with a model on the user's behalf. Each user message starts a turn. Within the turn the model may call tools (read a file, run tests, search), the harness executes them and sends the results back, and the model continues. A turn with $k$ tool steps is $k{+}1$ API requests, each carrying the whole conversation so far. An agentic turn of eighteen tool steps re-sends the conversation nineteen times.

\paragraph{The context window.} The conversation grows with every turn and tool step until it meets the model's context window, the ceiling on what one request can hold, input and output together: 1M tokens for Fable 5.1, Opus 5 and Sonnet 5, and 200k for Haiku 4.5~\citep{anthropic_context}. A request whose input alone exceeds the window is rejected. Before that point the harness compacts the conversation, replacing its history with a summary and starting its cache afresh~\citep{cc_caching,cc_envvars}, or the user starts a new one. A continuing conversation is therefore one whose context keeps filling, and every request in it re-reads more than the last until compaction resets it.

\paragraph{The cache.} Vendors bill re-sent context at a discount if the request shares a prefix with a recent request to the same model: the prompt cache. The prefix must match exactly and the cache belongs to one model, so sending the same conversation to another model recomputes everything~\citep{cc_caching,anthropic_caching}. Cache reads cost a fraction of the input price, cache writes carry a premium, and entries expire after minutes.

\paragraph{The price sheet.} On 21 September 2026 Anthropic listed its four models, dearest first, at \$10 and \$50 per million input and output tokens for Fable 5.1, \$5 and \$25 for Opus 5, \$2 and \$10 for Sonnet 5 and \$1 and \$5 for Haiku 4.5~\citep{anthropic_pricing}. Table~\ref{tab:regimes} gives the cache rules the three large vendors apply on top of their prices, and everything that follows turns on them.

\begin{table}[t]\centering\small
\caption{Prompt-cache price rules by vendor (Anthropic's of 21 September 2026, OpenAI's and Google's of 23 September 2026). Multipliers are of the model's input price.}\label{tab:regimes}
\begin{tabular}{@{}L{2.8cm}L{3.2cm}L{3.5cm}L{2.0cm}L{2.5cm}@{}}\toprule
Vendor & Cache read & Cache write & Lifetime & Minimum prefix \\\midrule
Anthropic~\citep{anthropic_pricing,anthropic_caching} & 0.025$\times$ (Fable 5.1), 0.1$\times$ others & 1.25$\times$ (5~min), 2$\times$ (1~h) & 5~min or 1~h & 512 to 4{,}096 tokens \\
OpenAI, GPT-5.6 and later~\citep{openai_caching,openai_pricing} & 0.1$\times$ & 1.25$\times$ & 30 min & 1{,}024 tokens \\
Google, Gemini~\citep{gemini_pricing} & 0.1$\times$ & input price plus a storage fee per hour (explicit cache) & not stated (implicit caching on by default) & 2{,}048 to 4{,}096 tokens \\
\bottomrule\end{tabular}
\end{table}

\paragraph{Layers, hooks, subagents.} The conversation is ordered as system prompt, project instructions, then messages, so a change deep in the conversation leaves the earlier layers cached~\citep{cc_caching}. Harnesses expose hooks that run at lifecycle events and can rewrite a tool call before it runs~\citep{cc_hooks}. They spawn subagents that start a fresh conversation with their own context~\citep{cc_subagents}, read instruction files (AGENTS.md, CLAUDE.md), load skills (SKILL.md folders) and connect tools through the Model Context Protocol (MCP).

\subsection{Model coupling is a design choice}\label{sec:coupling}
Table~\ref{tab:landscape} (Appendix~\ref{app:landscape}) grades twenty-one harnesses on the properties an enterprise inherits. Three facts stand out. First, openness and model coupling are independent axes. Gemini CLI is Apache-2.0 but talks only to Google, while Amp and Copilot CLI are proprietary but multi-model. Claude Code is proprietary and works with Anthropic's models only: its public repository holds issues and plugins but no source, and its documentation states that routing it to non-Claude models through a gateway is unsupported~\citep{cc_license,cc_gateway}. Second, where a default is documented, subagents inherit the parent's model (eight of the twenty-one), and the rest set the model per agent, leave it undocumented or have no subagents. Third, the harnesses of Chinese vendors, Qwen Code, DeepSeek Harness, Kimi Code, Trae Agent, ZCode and MiniMax Code, are MIT or Apache-2.0 and accept any OpenAI- or Anthropic-compatible endpoint, with CodeBuddy and iFlow CLI the closed exceptions. Model coupling is therefore a design choice of each vendor, not a property of the category.

%% file: sections/03_problem.tex
\section{Problem statement}\label{sec:problem}

Adopting a vendor's harness transfers four risks to the enterprise, and each is set by configuration an enterprise can read today (Table~\ref{tab:landscape}).

\subsection{Coupling ties the enterprise to one vendor's prices}
A harness bound to one vendor's models binds the enterprise to that vendor's prices, plans and limits. The binding is graded, not binary. Kiro accepts no other endpoint, Antigravity no other vendor's endpoint, and Claude Code no other vendor's models (a gateway can sit in front of it, for Claude models only). Cursor accepts a few vendors' keys, and Codex, OpenCode, pi, Goose, Aider, Cline and Copilot CLI accept an enterprise's own or third-party endpoints. Cloud marketplaces (Bedrock, Vertex, Foundry) change the invoice, not the model or its price structure~\citep{bedrock_pricing}. The exposure is not hypothetical. Cursor replaced request-based plans with usage-based pricing in June 2025 and apologised and refunded in July~\citep{cursor_pricing2025}, Anthropic announced weekly rate limits for its Pro and Max subscriptions in late July, citing demand for Claude Code~\citep{anthropic_limits2025}, and GitHub announced in April 2026 that Copilot moves to usage-based billing in June~\citep{copilot_billing2026}. Developers do not rank lock-in among their top concerns~\citep{jetbrains2025}, so it lands on procurement and the platform team.

\subsection{Enterprises pay for behaviour they cannot inspect}
Closed harness code means the enterprise cannot read what is sent on its behalf, audit prompt changes between versions, or verify how the cache, retries and subagents behave. The gap is real for Claude Code, Cursor, Kiro, Amp, Antigravity, CodeBuddy and iFlow CLI, partial for Copilot CLI (public repository, proprietary licence), and absent for the Apache-2.0 and MIT harnesses. Opacity compounds the pricing exposure: a bill set by a coupled harness is also a bill the enterprise cannot audit. It also goes largely unmanaged. Anthropic finds enterprise API use ``surprisingly price-insensitive''~\citep{aei2025sept}, so a saving has to come from the defaults, not from users reacting to prices.

\subsection{Connectors travel, the control plane does not}
A workforce split between two harnesses runs two skill bundles, two sets of connectors, two governance configurations and no shared telemetry. Three layers are already portable. MCP carries connectors in every surveyed harness except Aider and pi, which refuses it by design. AGENTS.md carries instructions: multi-vendor since August 2025, it is stewarded by the Linux Foundation and used by over 60,000 projects~\citep{agentsmd}. Agent Skills, an open standard since December 2025, carries procedures, with official adoption by OpenCode, GitHub Copilot, Gemini CLI, Cursor, Junie and Goose~\citep{agentskills}. What is not portable is exactly the control plane: model policy, telemetry, hook and permission semantics, subagent behaviour and billing.

\subsection{Novices get half the results for the same spend}\label{sec:skill}
The same harness produces very different results for different users. Anthropic's analysis of about 400,000 Claude Code sessions rates each session's user expertise from the precision of directions, what the user asks to be verified, and who corrects whom, and finds the rating task-specific and ``quite different from job title''~\citep{cc_usage}. Novice-rated sessions reach verified success 15\% of the time, against 28 to 33\% for intermediate users and above, and 19\% of them end abandoned, against 5 to 7\% for everyone else. A novice prompt sets off about five actions and 600 words of output, an expert prompt twelve actions and 3,200 words~\citep{cc_usage}. A randomised trial found experienced open-source developers 19\% slower with AI tools while they believed they were faster~\citep{metr2025}. Anthropic's own data shows higher-tenure users succeeding about 3 percentage points more often than others after controlling for the task, 10\% in relative terms~\citep{aei_mar26}, and DORA summarises the mechanism as ``AI doesn't fix a team; it amplifies what's already there''~\citep{dora2025}.

The cost consequence is direct. For a novice-rated user the same spend buys about half the verified successes, and a fifth of the sessions produce no outcome at all. A static instruction file does not close the gap on its own: repository context files do not raise an autonomous agent's success on well-specified benchmark tasks and add over 20\% to inference cost, while explicit instructions are followed and files help for non-standard practices~\citep{gloaguen2026agentsmd,khatri2026contextfiles}. The lever that reaches every user is the harness itself. Training reaches some users and scales with headcount, while harness configuration reaches every user at no marginal cost. The three expertise signals map onto features a harness can ship: scaffolds for precise requests, hooks that verify before a task completes, review gates, and a model policy that lowers the price of the waste that remains.

\subsection{One layer decides all four risks}
Every risk above is decided in the layer between the user and the model vendor: which model, which prompts, which subagents, which telemetry, which floors. Enterprises can own that layer today without owning the harness (Section~\ref{sec:actions}). Governance constrains how. Per-user cost and productivity metrics are the obvious by-product of a routing layer and the most dangerous one: in the EU, employee monitoring by AI falls under the AI Act's high-risk provisions and, in Germany, under works-council co-determination~\citep{eu_works_council}. The design in Section~\ref{sec:routing} therefore has no productivity axis and reports aggregates only. Data retention and refusal policies differ by model and exclude some workloads in some organisations, and the routing policy carries those floors. Cost is the risk with a price tag, and the next two sections price it.

%% file: sections/04_routing.tex
\section{Methodology}\label{sec:routing}

Routing inside a harness is a cache problem: the unit to price is the turn, a switch must earn back its cache writes, and a lower list price does not mean a lower session cost. The router and the emulation below are built on these three rules.

\subsection{Price the turn, not the request}
In a harness one user turn is $k{+}1$ API requests over a conversation that is cached on one model. That is the unit a router has to price, and it is where routing for chat products and routing for harnesses part ways. The routing literature chooses a model per request from an estimate of difficulty and treats each request as free to place~\citep{routellm2024,frugalgpt2023,llmrouterbench2026}, but in a harness the request carries the cached conversation. Writing $C$ for the conversation prefix, $D$ for the new tokens a turn adds, $O$ for its output and $k$ for its tool steps, a warm turn reads roughly $(k{+}1)C + (D{+}O)k/2$ cached tokens and writes $D + O$, because each request's output returns as input to the next (Appendix~\ref{app:method}). Moving the turn to another model turns the reads into a cold write of $C$ at the write premium and, when the user continues on the original model, a second cold write there.

\subsection{A mid-task switch must earn back its cache writes}\label{sec:payback}
Let $\Rh$ be the home model's cache-read price per token, $\Rt$ the target's read price, $\Wt$ the target's write price and $\Wh$ the home model's rewrite price. A mid-task switch that later returns pays only if the cheaper stretch lasts more than
\begin{equation}
n^{*} = \frac{\Wt + \Wh - \Rt - \Rh}{\Rh - \Rt}\ \text{requests},\qquad
n^{*}_{\text{no return}} = \frac{\Wt - \Rt}{\Rh - \Rt}.
\label{eq:payback}
\end{equation}
\begin{figure}[!b]\centering 
\includegraphics[width=0.62\linewidth]{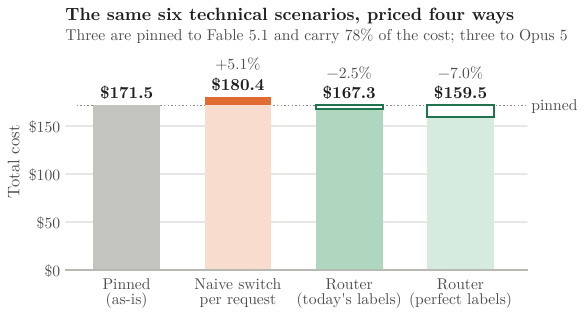}
\caption{Six technical scenarios of the catalogue, each counted once on the model it was written for; each bar is their total. The naive per-request switch pays a cold cache write on every move and redoes a turn the cheaper model was too weak for; the router runs small tasks in a separate lane. Sessions already on Fable 5.1 cannot gain from moving to the top model. Section~\ref{sec:case} finds 13 to 21\% of model spend by weighting each scenario by how often each segment runs it, on that segment's usual models.}\label{fig:switch}
\end{figure}
The return request rewrites the home cache instead of reading it, so it costs $\Wh - \Rh$ more than staying. On Anthropic's sheet, Fable 5.1 to Sonnet 5 gives $n^{*} = 291$ requests with return and 46 without, and Opus 5 to Sonnet 5 gives 27 and 8. The equation counts only the cached conversation, so these counts are the limit for long contexts. New tokens and output are cheaper on the target too, while the return also rewrites what the stretch added. With the turns of Figure~\ref{fig:crossover} and 50k to 200k tokens of context, a switch that returns pays after one to three turns from Opus 5 and after three to eleven turns from Fable 5.1, and one that never returns pays within its first turn. A naive switch can therefore pay on an Opus-pinned session after as little as one routine turn, and on a Fable-pinned session only over a longer run of them. On the catalogue's six technical scenarios, most of whose cost is pinned to Fable 5.1, the naive switch costs 5\% more than pinning and the router saves 2.5\% (Figure~\ref{fig:switch}); on the three pinned to Opus 5, the naive switch saves 8\%, as the payback rule allows, and the router 14\%. GitHub Copilot's documentation states the same observation operationally, routing ``along natural cache boundaries'' because switching mid-session ``has shown increased cost''~\citep{copilot_auto}. Equation~\ref{eq:payback} says when and by how much.

\subsection{The top model undercuts the next tier on long sessions}\label{sec:crossover}
\begin{figure}[!tb]\centering
\includegraphics[width=0.98\linewidth]{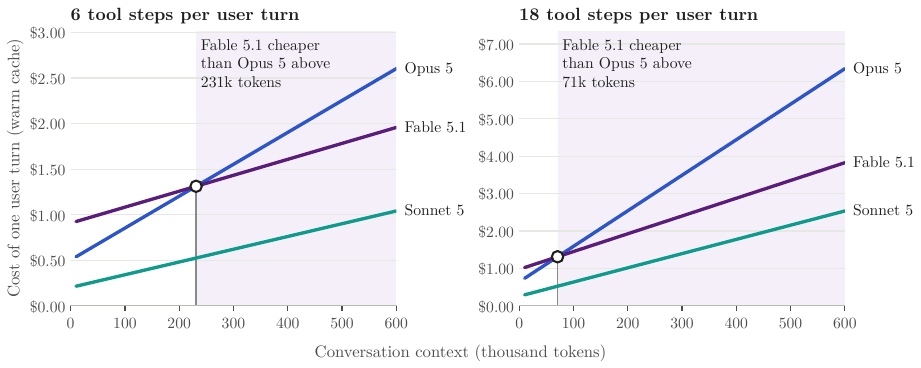}
\caption{Cost of one user turn with a warm cache against conversation context, for 6 and 18 tool steps per turn (40k new tokens and 6k output per turn; list prices of 21 September 2026). The break-even moves left as tool steps grow.}\label{fig:crossover}
\end{figure}

\begin{figure}[!tb]\centering
\includegraphics[width=0.98\linewidth]{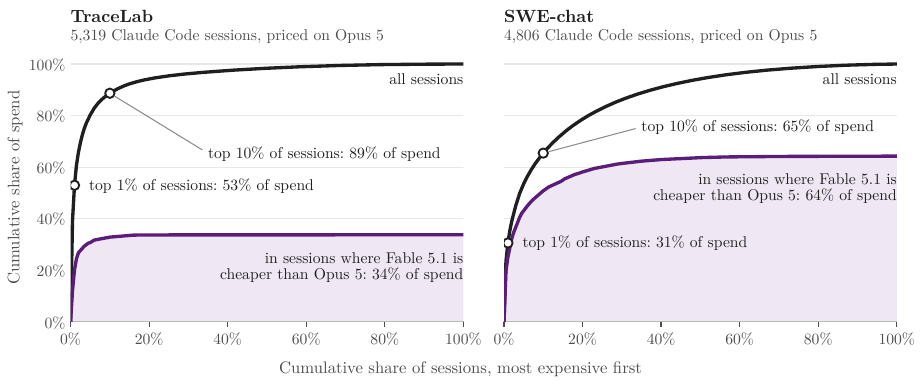}
\caption{Where real-session spend sits: the Claude Code sessions of TraceLab and SWE-chat priced on Opus 5 (list prices of 21 September 2026), most expensive first. A few long sessions carry most of the spend: the black curve counts all sessions, the purple curve only the sessions that cost less on Fable 5.1.}\label{fig:pareto}
\end{figure}

Cache reads dominate long agentic sessions, and vendors price them per model. On Anthropic's sheet of 21 September 2026, Fable 5.1 reads cached context at \$0.25 per million tokens against \$0.50 for Opus 5, while its input and output prices are double~\citep{anthropic_pricing}. Figure~\ref{fig:crossover} plots the cost of one warm user turn against context size. With eighteen tool steps per turn the highest-priced model becomes cheaper than Opus 5 above 71k tokens of context, and with six steps above 231k, both well inside the 1M-token window (Table~\ref{tab:breakeven}). A router that ranks models by list price gets long sessions exactly wrong. Real sessions bear this out. Repriced token for token on both models, the Claude Code sessions in two public corpora that cost \$100 or more on Opus 5 are cheaper on Fable 5.1 in 51\% (TraceLab) and 80\% (SWE-chat) of cases, and spend concentrates in exactly such sessions: the top 1\% of sessions carry 31 to 53\% of it (Figure~\ref{fig:pareto}; our repricing of~\citep{tracelab2026,swechat2026}). This one effect produces three quarters of the saving in the case study (Section~\ref{sec:decomp}).

\begin{center}\small
\captionof{table}{Context above which Fable 5.1 is cheaper than Opus 5 for one warm turn, by tool steps per turn (assumptions as in Figure~\ref{fig:crossover}).}\label{tab:breakeven}
\input{figures/tab_breakeven}
\end{center}

\FloatBarrier 
\subsection{Route without touching running work}
The rules follow from the arithmetic.
\begin{enumerate}[label=R\arabic*.]
\item Decide the model when the session is small, and never rewrite a running conversation's model.
\item Run bounded side requests in a \emph{separate} conversation seeded with a compact project ledger rather than the transcript, so the side lane writes a seed of about 20k tokens instead of the whole context.
\item After a side lane returns, refresh the home cache with one empty request, so the next home turn does not pay a full rewrite.
\item Append the side lane's result as a note; never edit earlier turns.
\item Branch only when the home context is large enough that the lane's setup is cheaper than re-reading in place.
\item Treat a new task as the moment to re-choose.
\item Choose by estimated session cost, not by tier.
\item Verify that any gateway forwards the cache markers.
\item Route subagents at launch: a subagent is a fresh conversation, so choosing its model costs nothing in cache terms, and fresh short conversations are write-heavy, where the top model is dearest.
\item Never move a correction below the tier of the turn it corrects.
\end{enumerate}
A side lane pays most where the context is largest: a one-line change inside an 805k-token session costs \$0.90 in place on Fable 5.1 and \$0.36 in a ledger side lane on Sonnet 5, the ledger and the home refresh included (Figure~\ref{fig:decision}). Five typical subagents cost \$5.09 when they inherit Fable 5.1 and \$1.32 on Sonnet 5, with the parent's own turn at \$0.25 either way, and Figure~\ref{fig:helpers} gives one subagent's cost by model and profile. Eight harnesses in Table~\ref{tab:landscape}, Claude Code among them, document that subagents inherit the parent's model by default. In Claude Code a pre-tool hook can rewrite the launch arguments, including the model, before the subagent starts~\citep{cc_hooks}.

\subsection{Label with a bring-your-own taxonomy, move only when sure}
Routing needs labels. We use a bring-your-own taxonomy of four axes (Appendix~\ref{app:taxonomy}): the action requested (thirteen labels from lookup to build and steer), the domain (twelve), an ordinal complexity (trivial, routine, complex, expert) and how much the request depends on context, from none to the conversation history, plus stakes and output-shape modifiers. Complexity maps to a capability tier, from Haiku 4.5 through Sonnet 5 (the mid tier) and Opus 5 to Fable 5.1. Domain sets governance floors: legal, finance and HR never go below the mid tier, and neither do agentic edits. Stakes can raise the tier but never above the user's own model. There is deliberately no productivity axis. The policy acts only when the classifier's probability that the needed tier is at or below the chosen one, and that the turn does not depend on history, both clear a threshold. Otherwise it \emph{holds} the turn on its current model, where it runs exactly as it would have without a router. A hold forgoes a saving, a misroute costs quality and rework, and Section~\ref{sec:holds} measures both. Corrections need care: ``no, that is wrong'' reads as trivial, but the steer label inherits the task's tier and the turn depends on the history, so the router holds it, and a separate classifier question asks of every turn whether it corrects the previous output (R10). Repeated corrections signal a task harder than it looked, and that is the moment to consider moving the task up, never down, at the price the payback rule gives.

\subsection{Jev classifies every prompt in under half a second}\label{sec:jev}
Every prompt waits for the routing decision, so the classifier sits on the user's critical path: a decision that takes two seconds is felt on every turn, one under half a second is not. That rules out a frontier model as the classifier, on latency and on cost, and rules in small purpose-built ones. Once a taxonomy bounds the answers to a small, enumerated set of labels, the classifier only has to score fixed options, a job that trades capacity the task never uses for latency.

The router uses Jev, a model of the class TypeSafe AI calls System One, released in September 2026~\citep{jev2026}. The name draws on the distinction between fast, intuitive System~1 thinking and slow, deliberate System~2 reasoning: Jev makes structured decisions and does not generate text. Its architecture explains its speed. A generative model writes its answer token by token, one forward pass per token, and a probability for each option needs several samples or access to its logits. Jev instead ingests the state once and evaluates every question against it in parallel: a new architecture with a parallel sampler generates all answers in a single query, and because the possible outputs are defined in advance, every answer is a typed value~\citep{jev2026,jevdocs}. A yes/no question returns a probability, a choice among up to 255 options returns a probability per option, and a score against a rubric returns a distribution over levels~\citep{jevcf}. Jev is trained with reinforcement learning for calibrated decisions (RLCD), so that higher confidence means higher accuracy~\citep{jev2026,jevdocs}. The vendor reports 70 to 500 milliseconds end to end, 40 to 200 times faster than frontier models of the same intelligence on decision-shaped queries, and about \$0.04 per million input tokens with output free~\citep{jev2026}. A request may carry 64k tokens (32k of state), the model is not trained on customer requests, and enterprises get zero data retention~\citep{jevdocs}.

We verified the interface live: the endpoint answered our four-axis questions in the documented shape (model string jev-1.13.0). Timed on all 109 catalogue prompts from a laptop over the public internet, a call took 250 milliseconds at the median and at most 366 once the connection was open. Only the first call, which opens the connection, took 0.7 seconds, so the gateway keeps its connection to the classifier open. An independent test on 60 hand-labelled cases measured 91.7\% accuracy, a calibration error near 0.07 and a median latency of about 420 milliseconds, inside the half-second budget, on a different task: rating the risk of agent tool calls~\citep{jevbench2026}. In the case study the classifier costs about \$179 a month for 1.3 million routed turns, and the router's other overhead, keeping caches warm and building ledgers, is \$5k a month (Section~\ref{sec:decomp}).

\begin{figure}[tbp]\centering
\includegraphics[width=0.99\linewidth]{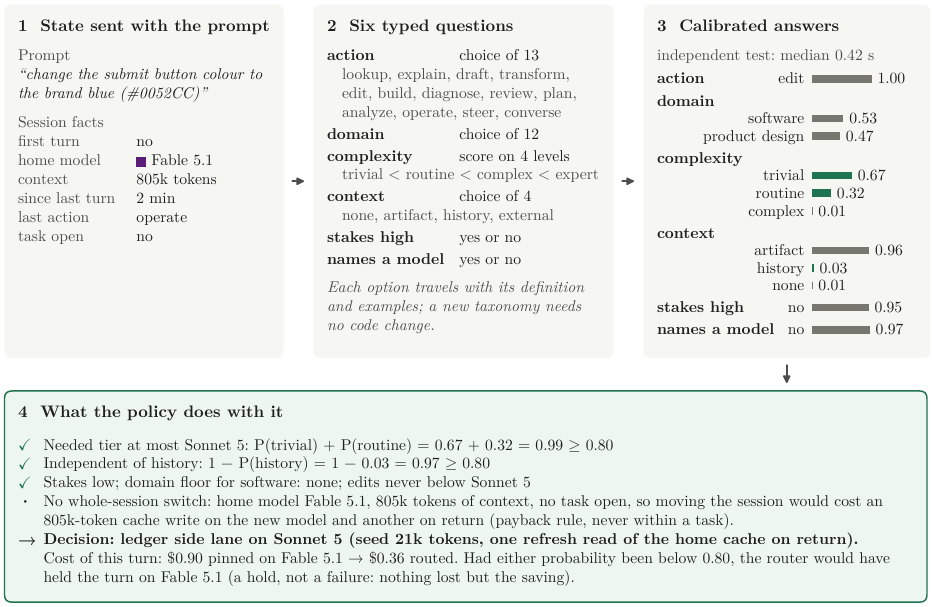}
\caption{One routing decision end to end: the one-line change inside an 805k-token session. The state and the six typed questions are what the classifier receives; the probabilities are its cached answer; the policy's arithmetic and the resulting side lane are shown at the bottom.}\label{fig:decision}
\end{figure}

\paragraph{Calibrated options beat a single label.} The per-option probabilities let the policy compute the probability that the \emph{decision} is right, that the tier the turn needs is at or below the cheaper model's and that the turn does not depend on history, and hold the turn when either is low. A single top label cannot express that, and a general model's stated confidence is not calibrated. Any classifier that returns calibrated per-option probabilities fast enough fits the same interface: small encoder classifiers of the kind vLLM's semantic router uses (a 307M-parameter encoder with several heads~\citep{vllmsr}), purpose-trained routing models such as Arch-Router (a 1.5B-parameter model aligned to a user-defined domain-and-action taxonomy~\citep{archrouter2025}), or an in-house encoder fine-tuned on the enterprise's own tagged corpus. Hosted classification means every prompt leaves the company, and Jev's weights and size are not published, nor is it offered for self-hosting. That is why the interface is classifier-agnostic and a self-hosted encoder is the stated fallback for restricted contexts. Figure~\ref{fig:decision} walks through one real decision from the catalogue: what the classifier received, what it answered, and what the policy did with it.

\subsection{Reasoning effort is a second dial}\label{sec:effort}
Model choice is one dial. The reasoning effort a model spends is another, and it does not always move cost the way the price sheet suggests. On bounded work, lower effort saves: in Anthropic's published runs on knowledge-work benchmarks, medium effort matched the default's accuracy at about 70 to 87\% of its cost, and ``below the model's ceiling, the highest effort levels pay for depth the task never uses''~\citep{anthropic_costopt}. On long-horizon, many-step work the direction can reverse. In ARC Prize's evaluation of GPT-6 Astra on ARC-AGI-3, high effort cost \$40.7k against \$48.1k for medium while scoring 54.8\% against 38.6\%, and maximum effort was the cheapest setting of all at \$26.1k and 62.7\%, ``because Astra solves games in fewer actions, reducing the total number of model calls and tokens''~\citep{arc_astra2026}. An observational study of 90 agentic coding runs found that raising effort from high to extra-high lifted first-try perfect runs from 28\% to 89\% and cut corrective prompts about fivefold~\citep{mehta2026effort}. Task shape decides, which is exactly what the taxonomy's complexity and output-shape axes describe, so effort is a natural second output of the same classifier and the next lever to route. Its metric is cost per completed task, not tokens per turn. One constraint carries over from the cache arithmetic: on most models each effort level has its own cache, so changing effort mid-conversation invalidates it like a model switch. Two exceptions exist on 21 September 2026: Fable 5.1 in Claude Code, and a per-message effort change, in beta, on Opus 5 and Fable 5.1 through a system message that leaves the prefix intact~\citep{cc_caching,anthropic_caching,anthropic_effort}. Effort routing therefore belongs at turn boundaries on models that keep the cache.

\subsection{Eight components, from gateway to governance floors}\label{sec:components}
The router consists of eight components. Our prototype implements components 2 to 6 and 8 as a decision engine and simulator, with the override rule of component 7. Components 1 and 7 connect the engine to live harnesses and their users.
\begin{enumerate}
\item \textbf{An enforcement point:} a gateway between every harness and its model endpoint for session-start decisions, cache-marker passthrough and measurement, plus harness hooks for in-session lanes and subagent launches.
\item \textbf{A classifier adapter} with the typed question set derived from the taxonomy (Section~\ref{sec:jev}).
\item \textbf{The bring-your-own taxonomy and policy:} tiers, floors, thresholds and the switches for each execution mode (Appendix~\ref{app:taxonomy}).
\item \textbf{Session state:} context size, cache age, whether a task is open, and the builder of the project ledger that seeds side lanes.
\item \textbf{The execution modes:} session-start choice; a detached request (a one-off call to a cheaper model without the conversation); a ledger side lane, followed by one empty request that refreshes the home cache; subagent launch routing; and, behind an evaluation gate, a rebase that restarts the home conversation from the ledger at a task boundary.
\item \textbf{Counterfactual accounting:} what each decision cost against what pinning would have cost, aggregated by unit, never by person.
\item \textbf{Transparency and override:} the user sees which model answered and can pin.
\item \textbf{Governance floors:} domain and data-class floors, a stakes ceiling, no productivity axis.
\end{enumerate}

\subsection{The emulation runs authored tasks with measured behaviour}\label{sec:emulation}
The emulation answers the question an enterprise faces before it has telemetry: what would routing save on our work, at our scale?

\paragraph{Tasks.} A scenario catalogue of 46 conversations with 109 authored turns across eleven job types (engineering, data, product, marketing, sales, legal, finance, HR, consulting, IT operations, general), each turn tagged with ground-truth labels and an activity profile: tool steps, tool-result tokens, output tokens, pauses. Selected consulting scenarios are shaped by the authors' own measured sessions: a slide deck built with a skill and a six-scout research run.

\paragraph{Measured behaviour.} Each conversation is expanded into 200 Monte Carlo variants that carry behaviour measured in about 10,000 real Claude Code sessions in two public corpora~\citep{swechat2026,tracelab2026}. After every task comes a chain of corrections drawn from SWE-chat's own pushback labels: 70\% of tasks draw none and 3\% five or more, and a correction does 5 to 13 tool calls, growing only slowly with the task. Pauses before follow-ups are drawn from SWE-chat (17\% longer than five minutes), subagents launch on 4\% of turns with tool work at TraceLab's rate (SWE-chat's 16\% as a sensitivity), and every turn carries the correction chain a misroute of it would cost. Engineering turn lengths follow the Claude Code sessions in TraceLab, whose upper tail they match (51 and 121 requests at the 75th and 90th percentile, weighted by requests, against 56 and 121). Prompt tokens come from a real tokenizer.

\paragraph{Classifier and pricing.} The live classifier labels every authored turn and every correction prompt from its text and structural session facts, and answers in a separate question whether a turn corrects the previous output (rule R10). The policy decides, and a cache-exact simulator prices pinned, naive-switch and routed execution from list prices and the vendors' caching rules: every tool step re-sends the conversation, with cache expiry, compaction, subagents and the side-lane overheads. Misroutes are charged as the correction chains users actually run: a turn moved below the tier its tags require, or away from history it needed, is redone on the home model and followed by further corrections, and a session started on too weak a model pays its first turn twice. Everything runs offline from the cached labels.

\paragraph{Population.} Results are scaled to a mixed population of 14 segments by function and usage intensity (power, regular and light engineers; heavy and light users in consulting and in data; one segment for each other function), each under its as-is model mix. ``Power'', ``regular'' and ``light'' are usage intensities (active days a month and sessions a day), not skill levels: a power user is an engineer who works in the harness all day. A seat runs 30 sessions a month on average, from 6 for light engineers to 72 for heavy consulting and research users. The as-is Fable share follows one rule, the top model for sessions that contain expert-level work: 20\% of power users' sessions, 15\% of regular users' and 4\% of light users', with the rest on Opus 5. Other segments use Opus 5, and Section~\ref{sec:sens} lets non-engineers pick Fable 5.1. The emulated engineers spend \$41 of model tokens per active day, between the median (\$15) and the mean (\$58) of TraceLab's heavy users (our pricing), and read 98.0\% of their input from cache, against 95 to 98\% in the two corpora.

%% file: figures/tab_breakeven.tex
\begin{tabular}{rl}
\toprule
Tool steps per user turn & Context above which Fable 5.1 is cheaper than Opus 5 \\
\midrule
3 & 422k \\
6 & 231k \\
12 & 114k \\
18 & 71k \\
30 & 34k \\
\bottomrule
\end{tabular}

%% file: sections/05_casestudy.tex
\section{Results}\label{sec:results}

\subsection{A 10,000-seat enterprise recovers 13 to 21\%}\label{sec:case}
\begin{figure}[!tb]\centering
\includegraphics[width=0.95\linewidth]{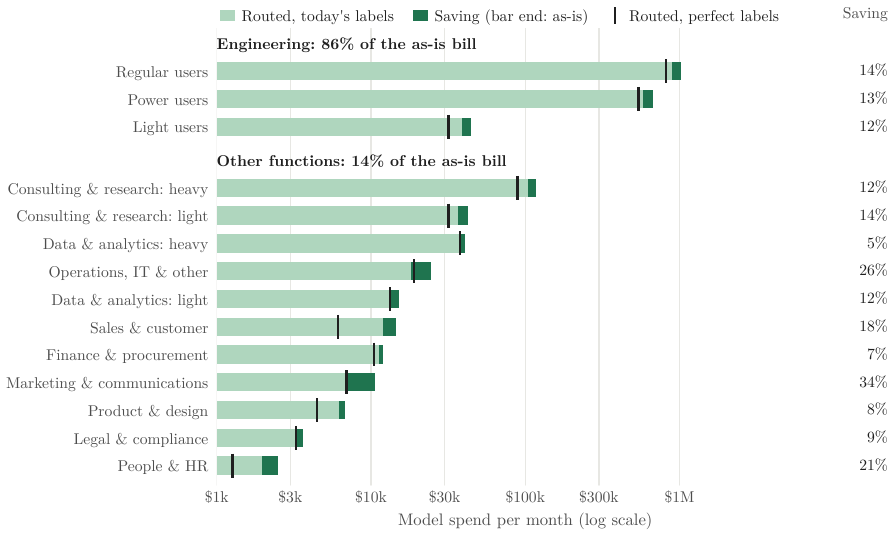}
\caption{Monthly model spend by population segment. Each bar ends at the as-is bill; the green tail is the saving with today's labels, the tick the routed bill with perfect labels. Engineering carries 86\% of the bill; non-engineering segments save 5 to 34\% of small bills.}\label{fig:population}
\end{figure}

As-is spend is \$2.03M a month, \$24.4M a year. Routed spend is \$1.76M with today's labels and \$1.61M with perfect labels: a saving of 13 to 21\%, \$3.3M to \$5.1M a year (Table~\ref{tab:summary}). By segment, engineers save 12 to 14\%, and the small non-engineering bills save 5 to 34\%, most in marketing and IT operations (Figure~\ref{fig:population}).

\begin{table}[!htb]\centering\small
\caption{Case-study summary (list prices of 21 September 2026; 10,000 seats; live classifier labels, cached).}\label{tab:summary}
\setlength{\tabcolsep}{5pt}\input{figures/tab_summary}
\end{table}

\begin{figure}[!tb]\centering
\includegraphics[width=0.8\linewidth]{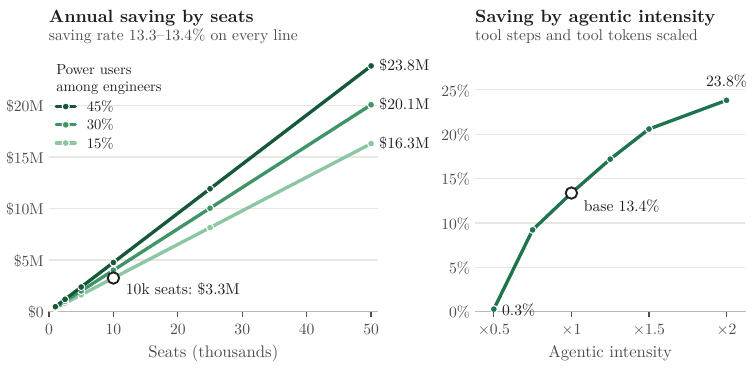}
\caption{How the saving scales. Left: annual saving against seats for three shares of power users among engineers, at a saving rate of 13.3 to 13.4\% on every line. Right: saving against agentic intensity, tool steps and tool-result tokens scaled together.}\label{fig:scale}
\end{figure}

\paragraph{Dollars scale with seats and power users.} Multiplied across an enterprise, the percentages become budget lines (Figure~\ref{fig:scale}). Dollars scale linearly with seats: the same population at 50,000 seats saves \$16.3M a year with today's labels, and Figure~\ref{fig:sizing} (Appendix~\ref{app:figures}) gives the saving for other seat counts and spend levels. Power users move the dollars but not the rate: raising their share from 15\% to 45\% of engineers lifts the as-is bill from \$2.03M to \$2.99M a month and the annual saving at 50,000 seats from \$16.3M to \$23.8M, at 13.3 to 13.4\% throughout. Agentic intensity moves the rate: the saving grows from 0.3\% at half today's tool steps and tokens per turn to 24\% at double, because the crossover grows with every re-read of a long context, and agentic work is moving towards the long end~\citep{codexdiff2026}. In dollars, the router earns most where sessions are tool-heavy and caches stay warm.

\subsection{Three quarters of the saving comes from the crossover}\label{sec:decomp}
\begin{figure}[!tb]\centering
\includegraphics[width=0.98\linewidth]{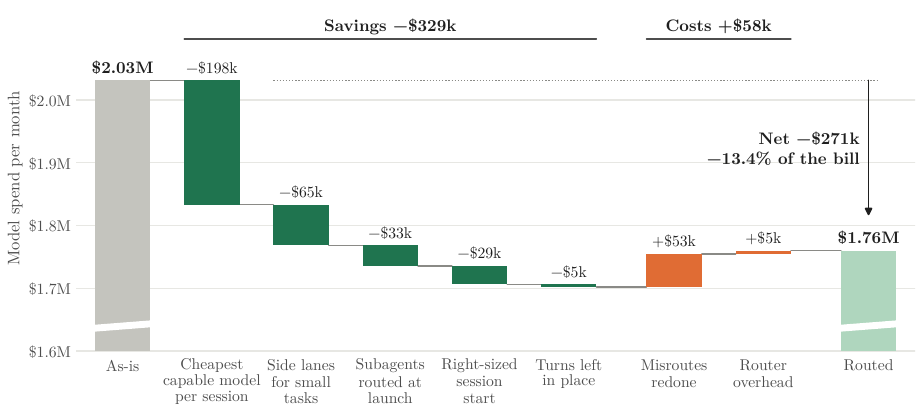}
\caption{Where the saving comes from (monthly, today's labels). The vertical axis is cut below the lowest step; the torn columns continue to zero.}\label{fig:mechanisms}
\end{figure}

Figure~\ref{fig:mechanisms} splits the saving: \$198k a month from choosing the cheapest capable model per session (the crossover of Section~\ref{sec:crossover}), \$65k from side lanes for small tasks, \$33k from routing subagents at launch, \$29k from right-sizing sessions at their start and \$5k on turns left in place, less \$53k of misroutes and their correction chains and \$5k of router overhead, almost all of it keeping the home conversation's cache warm across side lanes and building ledgers. The classifier calls themselves cost about \$179 a month.

\begin{figure}[!tb]\centering
\includegraphics[width=0.99\linewidth]{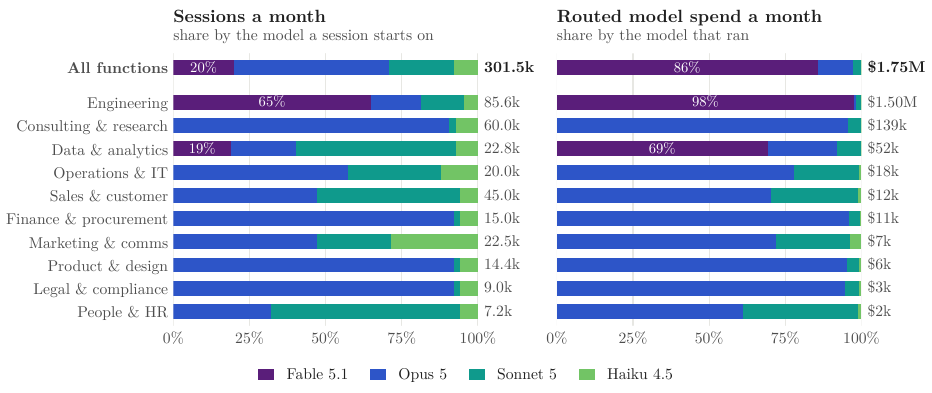}
\caption{Who runs on which model after routing (monthly, today's labels). Left: each function's sessions by the model a session starts on. Right: its routed model spend by the model that ran, with subagents, side lanes and redone misroutes, without router overhead. Totals at the end of each bar. As-is, every function outside engineering runs on Opus 5.}\label{fig:functions}
\end{figure}

\paragraph{Fable 5.1 runs the long builds.} Fable 5.1 takes 86\% of routed model spend but starts only a fifth of the sessions (Figure~\ref{fig:functions}): the long engineering and data builds that make up most of the bill. The sessions it takes from Opus 5 cost 13\% less on it. No session in marketing, sales, legal, HR, finance, product or consulting runs on Fable 5.1: Sonnet 5 is cheaper on every price line, so the router weighs Fable 5.1 only against Opus 5, and only for tool work. The choice is made at session start, before a session's length is known: half of the sessions moved to Fable 5.1 cost more on it, \$26k a month in all, against \$223k saved on the rest, \$197k net. Figure~\ref{fig:mechanisms} shows \$198k because it books these sessions' subagents and side lanes under their own mechanisms.

\subsection{Holds cost money, not quality}\label{sec:holds}
The classifier gets the action right 74\% of the time, the domain 80\%, the complexity level 63\% and history dependence 82\%. Among the turns the router moved, 12.5\% were false downgrades, moved below the tier their tags required. Corrections are a cost of their own: without them the as-is bill would be \$1.31M, so they make up more than a third of it. Corrections are also the most common reason people interrupt Claude Code, 32\% of interruptions in Anthropic's sample~\citep{anthropic_autonomy}. The classifier flagged every generated correction as one and the router moved none (rule R10). It also flagged 5\% of new tasks, which R10 then holds.

A turn is \emph{routable} when its ground-truth tags allow a cheaper model than the one it ran on. The router \emph{holds} a routable turn when the classifier's probability that the cheaper tier suffices, or that the turn does not depend on history, falls below the threshold: the turn stays on its current model and runs exactly as it would have without a router. A hold is not a failed decision, because nothing is lost but the saving. With today's labels the router held 42\% of routable turns. Of the \$150k a month between today's labels and perfect labels, two thirds (\$98k) are holds, mostly sessions that could have started on a cheaper model, and one third (\$53k) the correction chains of misroutes. It is a deliberate asymmetry: a wrong move costs quality and rework, a hold costs only money, so the policy is tuned to move only when sure. The 42\% comes from the classifier's complexity probabilities, which split between neighbouring levels, most often routine against complex, exactly where the taxonomy's level definitions are least crisp (63\% complexity accuracy, with the errors one level off). The remedy is therefore not a different classifier but sharper level definitions and examples in the taxonomy, tagged on the enterprise's real prompts, and a threshold set on that data. Section~\ref{sec:sens} shows the threshold trade-off.

\subsection{Pauses, the price sheet and tool use move the saving most}\label{sec:sens}
\begin{figure}[!tb]\centering
\includegraphics[width=\linewidth]{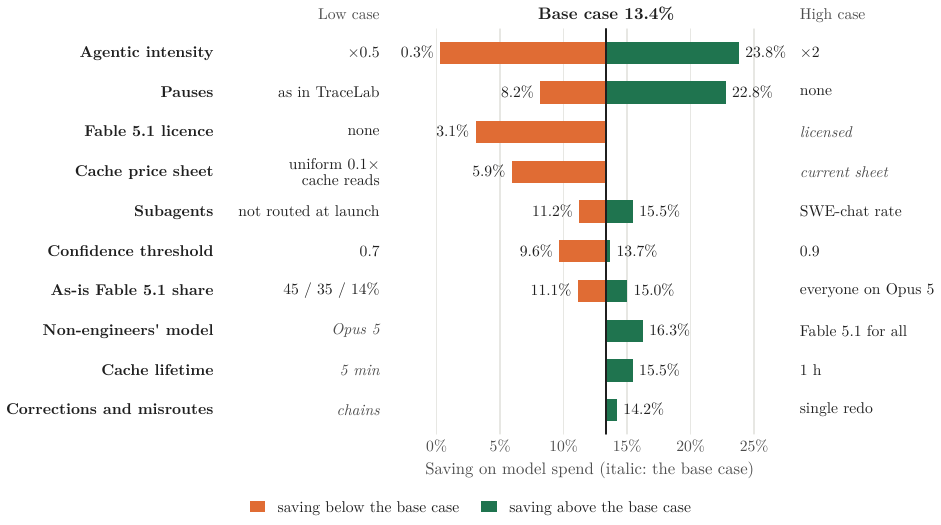}
\caption{One-at-a-time sensitivities of the population saving, one row per driver with its low and high case; italic case names mark the base (base 13.4\%; same labels and random draws). Orange bars run below the base, green bars above it. Table~\ref{tab:sensitivities} in Appendix~\ref{app:figures} lists every case.}\label{fig:tornado}
\end{figure}

Figure~\ref{fig:tornado} and Table~\ref{tab:sensitivities} (Appendix~\ref{app:figures}) vary one assumption at a time on the same labels and the same random draws. Four results matter for an enterprise reader. First, pauses. At the five-minute cache lifetime an API key gets, a user who returns after a pause finds the cache expired and pays for the whole conversation again, at the top model's higher write price. Without such pauses the saving would be 23\%, and with TraceLab's more pause-heavy users it is 8\%. A one-hour cache lifetime lowers the as-is bill and lifts the saving to 15.5\%, 16.4\% against today's bill: the cheapest default to change. Second, the price sheet. Under a uniform 0.1$\times$ cache-read regime with the same write premium, OpenAI's stated rule and Anthropic's sheet without the Fable discount, the as-is bill is 17\% higher (\$2.37M a month) and the saving falls to 6\%: the crossover lever disappears, and misroutes eat most of what side lanes and subagent routing save. Without a Fable 5.1 licence the saving falls to 3.1\%, because the crossover needs the top model. Third, agentic intensity (Section~\ref{sec:case}). Fourth, the classifier threshold (Figure~\ref{fig:thresholds}). Lowering it from 0.8 to 0.5 moves 46\% more turns, a fifth of them false downgrades whose correction chains cost more than the moves save, and the saving falls to 10\%. Raising it to 0.9 saves 13.7\% while holding half the routable turns. Moving more turns is not better: on this corpus the best tested threshold is 0.9, just ahead of 0.8. We keep 0.8 as the base, because choosing on this corpus would tune on the test set.

The behaviour sensitivities are smaller. Without corrections the bill is 36\% lower and the saving 13.8\%. Subagents at SWE-chat's rate lift the saving to 15.5\%, and without launch routing it falls to 11.2\%. Charging a misroute as a single redo would flatter the router by a point, and the as-is Fable share moves the saving between 14.7\% (3\% of engineering sessions, as observed in public data) and 11.1\% (Fable 5.1 for all complex or expert work). Outside engineering the base case keeps everyone on Opus 5, although in the authors' experience of enterprise rollouts many non-technical users pick the top model even for easy work. Anthropic's data shows the same coarse choice: paid Claude.ai users run 51\% of all their use on Opus, its most capable model class at the time, and only slightly more, 55\%, of computer and mathematical work~\citep{aei_mar26}. If non-engineers pick Fable 5.1 for half their sessions, the as-is bill rises 2.4\% and the saving to 14.8\%, and for all of them to 16.3\%. Their sessions are short chats, so the money stays in engineering's long sessions.

\subsection{Strong models of other vendors cost 0.12 to 0.80 times Opus 5}\label{sec:lockin}
\begin{figure}[!tb]\centering
\includegraphics[width=0.98\linewidth]{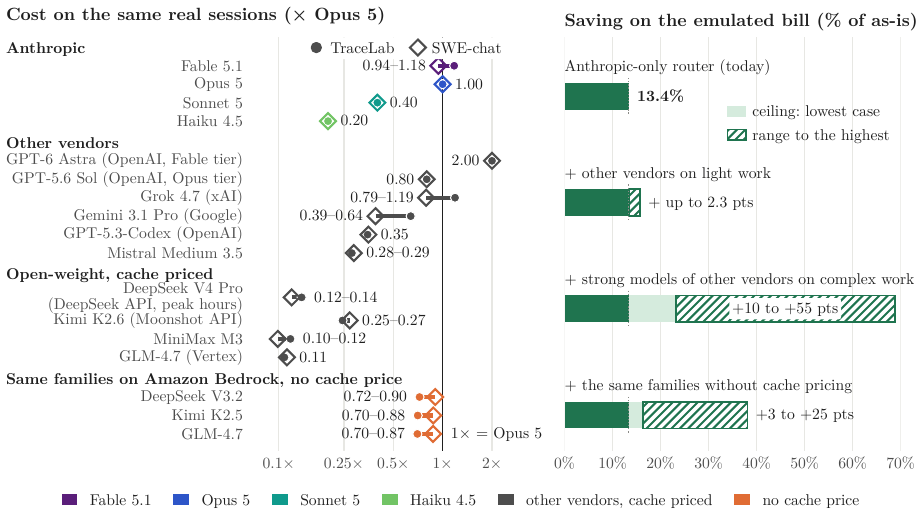}
\caption{The price of lock-in. Left: every Claude Code session in TraceLab and SWE-chat repriced token for token at each route's list prices (23 September 2026), as a multiple of the same sessions on Opus 5; each bar spans the two corpora. Anthropic's models keep their colours from Figure~\ref{fig:functions}; other vendors' routes are grey where they price cache reads and orange where they list no cache price. Right: the saving on the emulated bill when other vendors take only light work, or when strong models of other vendors take complex work at session start, with and without cache pricing. Ceilings: equal quality and the same token use on every model.}\label{fig:lockin}
\end{figure}

A router behind Claude Code chooses only among Claude models, because the harness does not support other vendors' models behind a gateway~\citep{cc_gateway}. A harness that accepts the enterprise's own endpoint (Table~\ref{tab:landscape}) can call them, and we priced that option on real sessions: every Claude Code session in TraceLab and SWE-chat, token for token, at each route's list prices of 23 September 2026~\citep{openai_pricing,gemini_pricing,mistral_pricing,xai_pricing,deepseek_pricing,moonshot_pricing,bedrock_pricing}. The cache-read price decides again (Figure~\ref{fig:lockin}). On routes that discount cache hits, strong models of other vendors cost 0.12 to 0.80 times Opus 5 on the same sessions: DeepSeek V4 Pro 0.12 to 0.14 at peak-hour prices, Kimi K2.6 0.25 to 0.27, Mistral Medium 3.5 0.28 to 0.29, GPT-5.3-Codex 0.35, Gemini 3.1 Pro 0.39 to 0.64 and GPT-5.6 Sol, OpenAI's counterpart to Opus 5, 0.80. DeepSeek, Kimi and GLM models on Amazon Bedrock, which lists no cache price for them, cost 0.70 to 0.90 times Opus 5. At the top tier OpenAI does not undercut Anthropic: GPT-6 Astra, its counterpart to Fable 5.1, costs 1.70 to 2.12 times Fable 5.1 on the same sessions, because its cache reads cost four times as much. xAI's Grok 4.7 costs 0.79 to 1.19 times Opus 5. On the emulated bill, other vendors that take only the light work the router already moves down add at most 2.3 points to the saving. Strong models of other vendors that take complex work at session start add 10 to 55 points to the 13\% on routes with cache pricing, and 3 to 25 without, at equal quality and token use.

%% file: figures/tab_summary.tex
\begin{tabular}{ll}
\toprule
Quantity & Value \\
\midrule
Scenarios / turns / job types & 46 / 109 / 11 \\
Classifier accuracy: action / domain / complexity / history & 74\% / 80\% / 63\% / 82\% \\
False downgrades among moved turns / held in place (of routable) & 12.5\% / 42\% \\
As-is spend, 10{,}000 seats & \$2,031k a month, \$24.4M a year \\
Routed, today's labels & \$1,760k a month ($-$13.4\%) \\
Routed, perfect labels & \$1,610k a month ($-$20.8\%) \\
\bottomrule
\end{tabular}

%% file: sections/06_actions.tex
\section{Recommendations}\label{sec:actions}

Two horizons, two decisions. The first can start this quarter, from within: put a router and a control plane behind the harnesses employees already use, between the harness and the model endpoint. The second is a product decision for the next one to two years: whether to own the harness itself.

\subsection{Now: put a router behind today's harnesses}
\paragraph{Own the control plane.} Put a gateway between every harness and its model endpoint (the harness's base-URL setting) and verify that it forwards cache markers and beta headers unchanged, because a gateway that strips them bills the whole conversation as uncached input on every turn~\citep{cc_caching}. Route at session start from the first prompt. Set a fleet default model for subagents today and route them per launch tomorrow. Export telemetry (token usage by type, session and model) and report aggregates only. Run a two-hour tagging workshop per function to build the routing taxonomy from that function's own prompts. The tags double as the policy review.

\paragraph{Ship one short bundle.} The router decides which model runs, and the bundle, shipped with the harness configuration, decides what the model can do and what every request carries. Ten items, in the order they pay off:
\begin{enumerate}
\item \textbf{Model policy:} the session-start rule, the subagent default, the domain and data floors, the user override.
\item \textbf{Skills as procedures:} how this enterprise deploys, reviews, names things and handles data, loaded on demand (the open Agent Skills format keeps only a description in context until a skill is used~\citep{agentskills}).
\item \textbf{Connectors:} the MCP servers a role needs, scoped by role, no more.
\item \textbf{Hooks:} tests and review before a task may complete, cache-marker checks on the gateway, the subagent launch rule.
\item \textbf{Permissions and data classes:} what tools may run unattended, which data never leaves for which model.
\item \textbf{The routing taxonomy} tagged by the function that uses it, with its thresholds.
\item \textbf{The project ledger format} that seeds side lanes and subagents instead of the transcript.
\item \textbf{Telemetry export and aggregate reporting}, wired to the counterfactual accounting.
\item \textbf{One short instruction file:} procedures and constraints, no repository description (Section~\ref{sec:skill}).
\item \textbf{Context defaults by role:} when old reasoning and tool results are cleared, when to compact, when to start a new conversation (below).
\end{enumerate}

\paragraph{Weigh the bundle.} Everything in it that is loaded on every request is re-read on every tool step. Table~\ref{tab:bundle} prices the always-loaded prefix for a ten-turn session of eighteen tool steps. The 55k row is real: one of the authors' own Claude Code sessions, read from the harness's context view, carried 5.9k tokens of system prompt, 15.8k of tool schemas, 15.4k of MCP tool schemas, 9.8k of skill descriptions, 6.4k describing 72 custom agents from a community plugin bundle, and 2.1k of memory before a single message. That configuration costs three to four times the lean default per session on every model, and it would cost more than ten times the default had the harness not deferred a further 104k tokens of tool schemas until needed. Two lessons follow. Harnesses already expose the accounting an enterprise needs to audit its bundle, per category and per tool: use it. And deferred loading, of tool schemas, skills and agent descriptions, is a cost lever in its own right, as the 104k deferred tokens show: cap what a role declares at start, load the rest when it is used, and measure.

\begin{table}[t]\centering\small
\caption{Cost per session of the always-loaded prefix (ten turns of eighteen tool steps, warm cache; list prices of 21 September 2026). Every token in the prefix is re-read on every request; the 55k row is a measured power-user configuration.}\label{tab:bundle}
\resizebox{\textwidth}{!}{\input{figures/tab_bundle_cost}}
\end{table}

\paragraph{Set context defaults by role.} Harnesses apply one context strategy to everyone. Opus 5, Sonnet 5 and Fable 5.1 keep every earlier turn's reasoning in context by default~\citep{api_context_editing}, and Claude Code compacts every conversation at one threshold~\citep{cc_envvars}. An engineer in the middle of a task may need that history, while a marketer revising a draft rarely needs the reasoning behind the previous version. Clearing old reasoning and tool results rewrites the cache from the cleared point on, so it follows the payback rule of Section~\ref{sec:payback}: on a warm cache it rarely pays, on an expired one it is nearly free, because the rewrite is due anyway. Expired caches are common. In two public corpora of Claude Code sessions, 17 to 27\% of user turns came after a pause of more than five minutes and 3 to 4\% after more than an hour (our analysis of~\citep{swechat2026,tracelab2026}), and Claude Code keeps the cache for an hour on a subscription but for five minutes on API keys and cloud providers~\citep{cc_caching}, which is how an enterprise gateway connects. A one-hour cache is therefore the first default to set.

For the returns that stay cold, Table~\ref{tab:coldreturn} prices the options. Ten thousand returns to a 300k-token history on Opus 5 cost \$18,750 just to re-send it, and pruning it to 30k first costs \$1,875. A summary by the same model costs more than the re-send on the return itself, because the summariser must read the whole history once, and pays back after two further requests. A summary by Sonnet 5 halves the cost of the return. A standard per role follows. Chat-shaped work: prune or summarise cheaply on every cold return, and start a new conversation per deliverable. Engineering: keep reasoning within a task, clear stale tool results in large batches, compact at task boundaries. Test answer quality on a pilot group before a standard rolls out, then set it centrally: an organisation's managed settings take precedence over every user's~\citep{cc_settings}. In Claude Code only the compaction threshold is such a setting, while clearing old reasoning and tool results needs the API's context-editing parameters, set by a gateway or an owned harness.

\begin{table}[t]\centering\small
\caption{Returning to a conversation whose cache has expired: 10,000 returns at 300k tokens of context on Opus 5 (list prices of 21 September 2026; 5-minute cache). Pruning clears old reasoning and tool results without a model call; a summary is written by the named model, which reads the full history uncached. The 30k is 15k of system prompt and tools plus a 15k-token summary. ``Each later request'' is the cache read of the history on every request after the return.}\label{tab:coldreturn}
\resizebox{\textwidth}{!}{\input{figures/tab_cold_return}}
\end{table}

\subsection{Later: decide on purpose whether to own the harness}
Table~\ref{tab:ladder} orders the options by what they buy and what they cost. Rungs one and two are this quarter's work and deliver the saving of Section~\ref{sec:case} on the harnesses people already use. Rung three is a different decision: an internal harness on a vendor SDK or an open-source base, with pluggable models, native lanes and one bundle for everyone. It is right for enterprises with regulated workloads that no vendor's retention terms fit, with their own models to serve, or with enough volume that a product team costs less than the dependency. It is wrong as a default, because vendors ship weekly and an open-source base moves the dependency to its maintainers. Decide it on purpose, after rung two has produced the numbers.

\paragraph{Keep the choice of models open.} The larger saving of Section~\ref{sec:lockin} turns on one decision, which models an enterprise accepts for its complex work, and only a harness that accepts the enterprise's own endpoint lets it take that decision. Most harnesses in Table~\ref{tab:landscape} accept one, and an owned harness always does.

\paragraph{Keep the record of what worked.} In our view an owned harness has a second, slower payoff. The taxonomy labels every prompt, and the router already detects corrections, so every exchange carries a signal of whether the answer stood or was corrected, the kind of feedback models are trained on. An enterprise that owns its harness and endpoints can turn this record into better procedures, routing and prompts, and later into tuned models. Few enterprises are set up to keep and govern such data today, and it is employee data, so it needs the same works-council and legal review as the policy (Table~\ref{tab:ladder}). Early work points the same way. Cursor retrains its completion model on accepted and rejected suggestions from over 400 million requests a day~\citep{cursor_tabrl}. RRSI evolves a harness's prompts, tools, skills and subagents around a frozen Claude Opus 4.8, gaining up to 6 points on the benchmarks it evolves against and up to 4.7 on unseen ones, and warns that unregularised self-evolution overfits~\citep{rrsi2026}. Dream-RSI replays an agent's own past runs offline to refine how it explores~\citep{dreamrsi2026}, and NVIDIA's guidance for long-running agents turns every production failure into a regression test, fixing prompts and tools first and weights last~\citep{nvidia_agentrl}. These are benchmark and vendor results, not measured enterprise gains, and they belong on the long-term side of the ladder.

\begin{table}[t]\centering\small
\caption{The sovereignty ladder.}\label{tab:ladder}
\begin{tabular}{@{}L{2.8cm}L{5.9cm}L{5.9cm}@{}}\toprule
Rung & What it buys & What it costs \\\midrule
0. Vendor defaults & Nothing to run & Every risk in Section~\ref{sec:problem}; the highest bill \\
1. Configured & The bundle above, model defaults, subagent default, permissions & Configuration per harness; no measurement \\
2. Controlled & Gateway, session-start routing, side lanes via hooks, telemetry, policy floors, taxonomy; the 13 to 21\% of Section~\ref{sec:case} & A small platform team; classifier hosting; works-council and legal review of the policy \\
3. Owned & An internal harness on a vendor SDK or an open-source base with pluggable models, native lanes, one bundle for everyone; other vendors' models (Figure~\ref{fig:lockin}) & A product team; keeping pace with weekly vendor releases; the dependency moves to maintainers \\
\bottomrule\end{tabular}
\end{table}

\paragraph{Ask vendors seven questions.} Does the harness forward cache markers through a gateway? Can it call an endpoint we operate? Can we set and enforce the model of subagents? Does it export token telemetry by type? What is the data retention of each model it offers? Can we pin a harness version and read its prompts? Which instruction, skill and connector formats does it read, and are they the open ones?

\paragraph{Measure cost per verified task.} Report it, and its spread across teams and harness configurations, in aggregate. Not tokens per user.

%% file: figures/tab_bundle_cost.tex
\begin{tabular}{lrrr}
\toprule
Always-loaded prefix per request & Fable 5.1 & Opus 5 & Sonnet 5 \\
\midrule
15k: lean default (system prompt and core tools) & \$0.90 & \$1.51 & \$0.60 \\
25k: default plus a 10k instruction and skills bundle & \$1.49 & \$2.52 & \$1.01 \\
55k: observed power-user configuration with plugins & \$3.29 & \$5.54 & \$2.22 \\
159k: the same if deferred tool schemas were loaded eagerly & \$9.50 & \$16.02 & \$6.41 \\
\bottomrule
\end{tabular}

%% file: figures/tab_cold_return.tex
\begin{tabular}{lrrr}
\toprule
10{,}000 returns at 300k context & On the return & Each later request & Return vs re-send \\
\midrule
Re-send the full history (default) & \$18,750 & \$1,500 & +0\% \\
Prune to 30k first (clear old reasoning and tool results) & \$1,875 & \$150 & -90\% \\
Compact to 30k, summary by Opus 5 & \$20,625 & \$150 & +10\% \\
Compact to 30k, summary by Sonnet 5 & \$9,375 & \$150 & -50\% \\
\bottomrule
\end{tabular}

%% file: sections/07_related.tex
\section{Related work}\label{sec:related}
\paragraph{Routing and cascades.} We price the vendor-side prompt cache of an agentic turn and derive when a switch pays, where the routing literature prices single requests. RouteLLM learns win-prediction routers from preference data~\citep{routellm2024}, FrugalGPT cascades models by confidence~\citep{frugalgpt2023}, and recent benchmarks and analyses score routers on independent queries~\citep{llmrouterbench2026,routingcollapse2026,ucci2026}, the largest on over 400,000 instances~\citep{llmrouterbench2026}. Multi-turn formulations treat routing as a budgeted sequential decision~\citep{seqroute2026}, and agent-serving simulators test policies that keep a program's successive turns on the same engine~\citep{agentservesim2026}. Neither prices a vendor's prompt cache: the first charges abstract per-token costs, the second works inside the serving engine.
\paragraph{Routers in harnesses.} Our payback rule explains what shipped routers do by rule of thumb. Copilot's auto mode routes along cache boundaries and states that mid-session switching raised cost~\citep{copilot_auto}. OpenRouter's sticky routing activates only when a provider's cache reads are cheaper than fresh input~\citep{openrouter_sticky}, and the open-source claude-code-router selects by static keys and has met the cache-prefix problem in the wild~\citep{ccr}. The concept of routing is not new, and independent work applied Jev to it within days of the model's release: those routers fall back to a safe tier when the classifier is unsure, and one refuses to downgrade long conversations to protect the cache~\citep{jevrouter_garg,jevrouter_gist}. OpenAI's GPT-5 router~\citep{gpt5router2025} met a user backlash over lost control~\citep{fortune_gpt5backlash2025}, which is why our design shows the model and allows a pin.
\paragraph{Caching for agents.} The nearest academic neighbour evaluates caching strategies for long-horizon agentic tasks across vendors without model switching~\citep{dontbreakcache2026}, and Codex's exact-prefix cache has been measured under model and effort changes~\citep{zentrik2026}. We take the cache as given and ask which model should own it.
\paragraph{Developer productivity and skill.} Our skill argument rests on Anthropic's session-level expertise analysis~\citep{cc_usage,aei_mar26}, the METR trial~\citep{metr2025,metr2026update} and DORA~\citep{dora2025}. The context-file ablations~\citep{gloaguen2026agentsmd,khatri2026contextfiles} bound what static files do for well-specified tasks and leave the unskilled-user case open (Section~\ref{sec:skill}). The difficulty distribution of coding-agent tasks~\citep{codexdiff2026} bounds how much work can move to cheaper models at all.
\paragraph{Governance.} Employee-monitoring constraints under the EU AI Act and co-determination law~\citep{eu_works_council} shaped the absence of a productivity axis.

%% file: sections/08_limitations.tex
\section{Limitations}\label{sec:limits}

The results rest on an emulation, two public corpora and one vendor's list prices. Every input below can be replaced with an enterprise's own, and the bring-your-own design exists for that.

\begin{itemize}
\item \textbf{Synthetic scenarios.} The 46 conversations and 109 turns are synthetic. The authors wrote them from their experience of leading teams and watching how people work with assistants and agents, and shaped selected scenarios on real sessions of their own. Measured behaviour perturbs them but adds no new tasks, and job types without a public corpus borrow the developers' behaviour on turns without tool work. Their ground-truth tags come from the authors alone, and future work should rest on real, independently tagged prompts wherever an enterprise can provide them.
\item \textbf{A taxonomy still in development.} The routing taxonomy is not final. Its complexity levels are its least crisp part and cause most holds, which is why every result is a band from today's labels to perfect labels. It has no ablations and no inter-rater agreement yet. The classifier's accuracy on a given team is known only once that team's prompts are tagged by two independent raters with their agreement reported.
\item \textbf{Session shape.} The engineering scenarios under-represent the lightest turns: a quarter of real requests sit in turns of fewer than seven requests, against 3.8\% of ours. They also re-read more of their input from cache (98.0\%, against 95.2\% in TraceLab and 97.6\% in SWE-chat), which favours the crossover. The share of long sessions that are cheaper on Fable 5.1 (51 to 80\%) rests on the 102 TraceLab and 30 SWE-chat sessions that cost \$100 or more. Repricing real sessions bounds the crossover at 6.0\% (TraceLab) to 12.6\% (SWE-chat) of an all-Opus bill, against 11.2\% of the Opus spend in the emulation. Sessions shaped like TraceLab's would bring the 13\% to about 9\%, in line with its pause sensitivity (8.2\%).
\item \textbf{Usage intensity and the dollars.} The emulated engineers spend \$41 per active day: between the median (\$15) and the mean (\$58) of TraceLab's heavy users, and three times the vendor's reported average of about \$13, with 90\% of users below \$30~\citep{cc_costs}. Volume scales the dollars and leaves each segment's percentage unchanged: at \$13 per engineer-day the bill is \$10.1M a year and the saving \$1.3M to \$2.2M. The blended rate moves to 13.3 to 21.3\% because engineering's share of the bill falls. Shorter sessions lower the percentage as well (Figure~\ref{fig:scale}).
\item \textbf{Scale and the cost of building a router.} The saving grows with seats (Figure~\ref{fig:scale}), and the cost of building and running a router does not. For a small organisation the saving may not pay for an engineering team to build one. Standard routers that need no engineering of their own are likely to appear and would remove that threshold.
\item \textbf{Model choice outside engineering.} Non-engineers start on Opus 5 in the base case. That many pick Fable 5.1 even for easy work is the authors' experience of enterprise rollouts, not a measurement, and it enters only as a sensitivity (14.8\% if they pick it for half their sessions, 16.3\% for all).
\item \textbf{Two corpora, developers only.} About 10,000 Claude Code sessions from SWE-chat and TraceLab~\citep{swechat2026,tracelab2026}. They disagree on subagent use by a factor of four per tool turn and on pauses by a factor of 1.6.
\item \textbf{List prices, no contract discounts.} Prices are the list prices of one vendor family on 21 September 2026. Enterprise contracts are not deducted, so the dollar totals may be lower. A discount that applies to every model lowers every dollar figure by its share and leaves the percentages unchanged, while discounts that differ by model move the percentages as well.
\item \textbf{Prices move fast.} Providers change prices and price structures often and at short notice, and every change moves the dollars and can move the percentages. Under a uniform 0.1$\times$ cache-read rule the crossover disappears and the saving falls to 6\% (Table~\ref{tab:sensitivities}). What transfers is the method: price every candidate on the enterprise's own sessions at the day's prices.
\item \textbf{Other vendors' models.} The lock-in ceilings (Figure~\ref{fig:lockin}) reprice real sessions at list prices. They assume equal quality and the same token use on every model, no cross-vendor cache events inside a session, and data terms that allow the route. DeepSeek is priced at peak hours, and its off-peak prices are half. OpenAI's models are priced at their short-context rates, because its page does not state where the long-context rates begin.
\item \textbf{The perfect-label ceiling.} Perfect labels know each turn's tier and treat capability as a step: a model at the needed tier always succeeds, one below always fails. Real difficulty is only partly visible in the prompt and real capability is a probability, so 21\% is a ceiling rather than a target.
\item \textbf{Classifier figures.} Accuracy and calibration are the vendor's figures, apart from one independent test on 60 hand-labelled cases~\citep{jevbench2026}. The latency is our own measurement from one laptop over the public internet. A hosted classifier sends every prompt out of the company, which the self-hosted fallback avoids.
\item \textbf{Prototype scope.} The decision engine, the policy and the simulator are implemented. The enforcement point (gateway and hooks) and the user's view of the routing are specified but not built, so the case study runs every decision in the simulator, not behind a live harness.
\item \textbf{Corrections move with the models.} Correction rates come from public sessions of 2026. Anthropic reports human interventions per Claude Code session falling from 5.4 to 3.3 as its models improved~\citep{anthropic_autonomy}, so later models may shrink the correction share of the bill and the cost of a misroute.
\item \textbf{Static users.} Behaviour does not react to the router: nobody changes how they work because a router exists.
\item \textbf{Levers outside the emulation.} Routing of reasoning effort (Section~\ref{sec:effort}), the quality effect of clearing old reasoning, for which Claude Code has no setting, and whether procedures and skills lift unskilled users on under-specified requests. The context-file ablations leave the last open, because they have no human in the loop and use well-specified tasks (Section~\ref{sec:skill}).
\end{itemize}

%% file: sections/09_conclusion.tex
\section{Conclusion}\label{sec:conclusion}
An enterprise's AI bill is set in two places: the model's price sheet and the harness. The price sheet lists the rates, and the harness picks which rate applies to every request and how many tokens are bought at it. It also decides governance and dependence. Enterprises that negotiate only the price sheet leave the harness, and with it much of the bill, to their vendor's defaults.

Routing is the most direct lever the harness offers, and inside a harness it obeys the cache. We gave the payback rule for mid-task switches, showed that for long sessions the highest-priced model costs less than the next tier and confirmed it on real sessions, and built a router that acts only where no running conversation rebuilds its cache, with Jev labelling every prompt against a bring-your-own taxonomy in under half a second. At 10,000 seats it recovers 13 to 21\% of model spend, \$3.3M to \$5.1M a year. Pauses decide as much as the classifier, because an expired cache turns every return into a full rewrite, which makes the cache lifetime part of the lever. Corrections make up more than a third of the bill and make every misroute expensive, so the router moves work only when it is sure. The harness also decides which vendors are in reach: on the same real sessions, strong models of other vendors cost 0.12 to 0.80 times Opus 5. Enterprises can own the control plane around their harnesses from within, starting this quarter, and decide about owning the harness on purpose rather than by default.

%% file: sections/A_taxonomy.tex
\section{The routing taxonomy}\label{app:taxonomy}
The taxonomy is bring-your-own: labels, descriptions and examples live in a configuration file that is sent to the classifier with every request and can be edited per enterprise; only the four context values are fixed, because the execution modes depend on their meaning. The default used in the case study has four axes and two modifiers.

\paragraph{Action (13, domain-neutral verbs).} lookup, explain, draft, transform, edit, build, diagnose, review, plan, analyze, operate, steer (a short control message inside an ongoing task, which inherits the task's tier), converse.

\paragraph{Domain (12).} software, data analytics, product and design, marketing and communications, sales and customer, legal and compliance, finance and procurement, people and HR, operations and IT, research and strategy, general or personal, unknown. Domain drives policy floors and the expected session shape, not difficulty.

\paragraph{Complexity (4, ordinal).} trivial (one step, one fact or a one-line change; under a minute of skilled work), routine (well specified and bounded; minutes to a quarter of an hour), complex (multi-step, needs synthesis or judgement; a quarter of an hour to hours), expert (novel, ambiguous or high-stakes reasoning; long-horizon agentic work where errors compound). Complexity maps to the capability tiers fast, balanced, strong and frontier.

\paragraph{Context (4, fixed).} none (the prompt alone suffices), artifact (needs the current working set but not the conversation's reasoning; reconstructible from a project ledger plus the files), history (needs prior reasoning, decisions or unfinished state from this conversation), external (needs information outside the session).

\paragraph{Modifiers.} stakes (a wrong answer has material consequences; may raise the tier but never above the user's own model) and output shape (short, medium, long; used for cost estimates).

\paragraph{Decision.} The policy computes, from the classifier's probability vectors, the probability that the tier the turn needs is at or below the tier of the candidate model and the probability that the turn does not depend on history. Both must clear a threshold (0.8 in the case study) for the router to act. The action steer, task-opening actions and turns in a task that is still open stay on the home model, and legal, finance and HR never go below the balanced tier, nor do agentic edits.

%% file: sections/B_method.tex
\section{Cost model and constants}\label{app:method}
\paragraph{Per-request accounting.} Each API request is billed in four categories: uncached input, cache read, cache write (5-minute or 1-hour) and output. For a user turn of $k$ tool steps ($k{+}1$ requests) at prefix $C$ with $D$ new tokens and output $O$ spread over the requests, each request's output is part of the next request's prompt, so it is written once and re-read like the new tokens. A warm cache, which holds all of $C$ but the output of the previous turn's last request, gives reads $(k{+}1)C + Dk/2 + O(k{-}2)/2$ and writes $D + O$; a cold cache writes $C + D + Ok/(k{+}1)$ and reads $kC + Dk/2 + Ok(k{-}1)/(2k{+}2)$. The simulator tracks per-model cache state, time since last touch against the lifetime, compaction at 85\% of the window (a 20k-token summary replaces the history), and charges the side lane's ledger build, its cache write and the one-request home refresh.
\paragraph{Misroutes.} A moved turn whose ground-truth complexity requires a higher tier than the model it ran on is redone on the home model; a turn moved away from history it needed likewise; a session started on too weak a model pays its first turn twice. The naive-switch baseline is charged by the same rule.
\paragraph{Subagents.} One subagent with $k$ tool steps, system prompt plus brief $P$ tokens, $t$ tool-result tokens per step, $o$ output tokens per step and a final report $r$ costs reads $kP + (t{+}o)\,k(k-1)/2$, writes $P + k(t{+}o)$ and output $ko + r$, since each step's output is re-sent with its tool result. Profiles: light (5 steps, 2k tokens per step), typical (12 steps, 3k) and heavy (60 steps, 2.5k), with $P = 16{,}200$.
\paragraph{Measured behaviour.} Every scenario is expanded into 200 variants (seed fixed; one random stream per behaviour and turn position, so a sensitivity changes one behaviour only), each carrying 1/200 of the scenario's weight. After every task, $k$ corrections are drawn from SWE-chat's distribution for tasks with or without tool work (probabilities for $k = 0$ to 12, a geometric tail above). A correction has 30 prompt tokens and the measured tool calls for its task's size (4.9, 7.0, 8.4, 10.5, 13.3 for tasks of 0, 1--5, 6--20, 21--60 and more tool calls), with the corrected turn's tokens per step. Pauses before follow-ups follow SWE-chat's quantiles, a turn with tool work launches subagents with probability 0.039 and the measured fan-out, and every turn carries a misroute chain drawn from the corrections distribution conditioned on at least one.
\paragraph{Constants.} System prompt and tools 15k tokens, ledger 6k, detached system prompt 1.5k, reintegration note 80. Cache lifetime 5 minutes (API default) or 1 hour (sensitivity). Each classifier call carries the prompt plus 350 tokens of session facts and the taxonomy's question set of 1,812 tokens, and rule R10 asks its correction question (136 tokens) in a second call with the same state, at \$0.042 per million. Prompt tokens are counted with the o200k\_base tokenizer scaled by 1.25, because the vendor's token-counting endpoint was unavailable. Prices are the list prices of 21 September 2026 (Section~\ref{sec:mechanics} and Table~\ref{tab:regimes}), and later releases are out of scope. Population: 10,000 seats in 14 segments by function and usage intensity, 301k sessions a month, engineering 86\% of spend.
\paragraph{Reproduction.} Every number, figure and table is produced offline by the authors' scripts (emulation, figures, worked examples, corpus calibration and the real-session repricing) from the catalogue, the cached classifier labels and the two public corpora; an audit script recomputes every printed number from the same files. The code and the catalogue are not released; Appendix~\ref{app:taxonomy} gives the taxonomy and policy and this appendix the cost model, so the method can be rebuilt with an enterprise's own inputs.

%% file: sections/C_landscape.tex
\begin{landscape}
\section{The harness landscape}\label{app:landscape}
Twenty-one agentic coding harnesses graded on the properties an enterprise inherits (Section~\ref{sec:coupling}).

\begin{center}\small
\captionof{table}{Coupling matrix of twenty-one agentic coding harnesses, from licence files and official configuration documents (September 2026). ``Own endpoint'': whether an enterprise can point the harness at its own or a third-party model server with other vendors' models (a gateway in front of the vendor's own models does not count); n/d: not documented in official sources.}\label{tab:landscape}
\resizebox{\linewidth}{!}{%
\begin{tabular}{@{}lllllll@{}}\toprule
Harness & Source & Model coupling & Own endpoint & Subagent model & Skills / MCP & Telemetry \\\midrule
Claude Code~\citep{cc_license} & proprietary, native binary & Anthropic only & no & field, inherits & yes / yes & OpenTelemetry \\
Codex~\citep{codex_cli} & Apache-2.0 & OpenAI default, custom providers & yes & config, per agent & yes / client & opt-in OTel \\
Gemini CLI~\citep{gemini_cli} & Apache-2.0 & Google only & no & field, inherits & yes / yes & OpenTelemetry \\
Antigravity~\citep{antigravity} & proprietary (Google terms) & Gemini; Claude, GPT-OSS on consumer plans & no (SDK: local models) & tier field, inherits & yes / yes & to Google; own cloud project on Enterprise \\
OpenCode~\citep{opencode} & MIT & 75+ providers & any & per agent & yes / yes & dropped in v2 \\
pi~\citep{pi_agent} & MIT & any supported API & yes & no subagents & yes / refused & none \\
Goose~\citep{goose} & Apache-2.0 & 15+ providers & yes & env, recipe & yes / yes & OTLP \\
Aider~\citep{aider} & Apache-2.0 & any via LiteLLM & yes & no subagents & no / no & none \\
Cline~\citep{cline} & Apache-2.0 & multi, any compatible & yes & inherits & yes / yes & enterprise \\
Cursor~\citep{cursor_cli} & closed & own + a few vendors' keys & no & field, inherits & yes / yes & enterprise \\
Copilot CLI~\citep{copilot_cli} & public, proprietary & Copilot + own keys & yes & field, inherits & yes / yes & announced \\
Kiro~\citep{kiro} & proprietary & Bedrock only & no & field & yes / yes & enterprise \\
Amp~\citep{amp} & closed & multi via routing & no & per mode & yes / yes & analytics API \\
Qwen Code~\citep{qwen_code} & Apache-2.0 (Gemini CLI fork) & Qwen default, OpenAI/Anthropic/local & yes & field, inherits & yes / yes & OTel, off by default \\
DeepSeek Harness~\citep{deepseek_harness} & MIT & model-agnostic (OpenAI, Anthropic protocols) & yes & delegate providers & plugins / plugins & third-party plugin \\
Kimi Code~\citep{kimi_code} & MIT & Kimi default, Anthropic/OpenAI/Google & yes & inherits; experimental pool & marketplace / yes & n/d \\
Trae Agent~\citep{trae_agent} & MIT & multi incl.\ Doubao, Ollama & yes & no subagents & no / yes & none \\
ZCode~\citep{zcode} & Apache-2.0 & GLM default, tool-agnostic & config override & n/d & yes / yes & n/d \\
MiniMax Code~\citep{minimax_code} & MIT & MiniMax default, OpenAI/Anthropic formats & yes & subagents, no override found & marketplace / yes & n/d \\
CodeBuddy~\citep{codebuddy} & closed, npm & Tencent models, custom endpoint & yes & per-agent overrides & no / yes & none found \\
iFlow CLI~\citep{iflow_cli} & closed, no licence file & Qwen, Kimi, DeepSeek; custom base URL & yes & agents, model n/d & no / marketplace & n/d \\
\bottomrule\end{tabular}}
\end{center}
\end{landscape}

%% file: sections/D_figures.tex
\section{Supplementary figures and tables}\label{app:figures}

\begin{figure}[!htb]\centering
\includegraphics[width=0.8\linewidth]{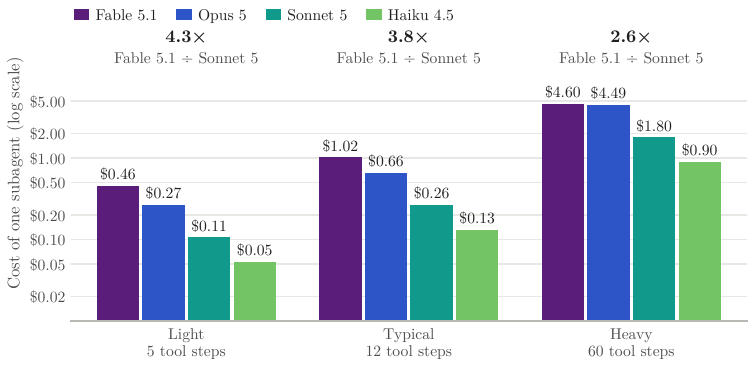}
\caption{Cost of one subagent by model and activity profile (15k-token system prompt, 1.2k-token brief; tool results and output as in Appendix~\ref{app:method}). Subagents write their whole context once and read it a dozen times; the gap between the top model and the mid tier is largest for short subagents and narrows for very long ones.}\label{fig:helpers}
\end{figure}

\begin{figure}[!htb]\centering
\includegraphics[width=0.7\linewidth]{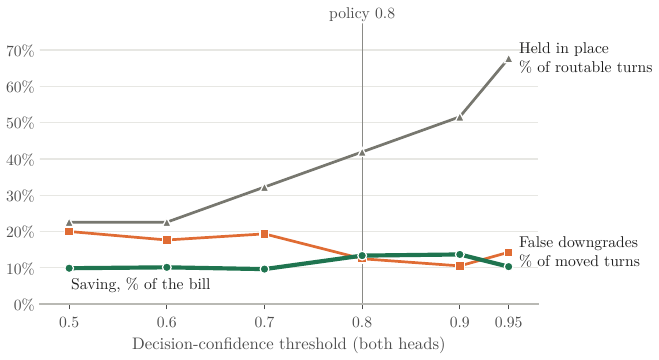}
\caption{Saving, false-downgrade rate and hold rate against the decision-confidence threshold, applied to both of the classifier's decision questions: the tier a turn needs and its dependence on the history.}\label{fig:thresholds}
\end{figure}

\begin{table}[!htb]\centering\small
\caption{Sensitivities (population, monthly, same labels).}\label{tab:sensitivities}
\resizebox{\textwidth}{!}{\input{figures/tab_sensitivities}}
\end{table}

\begin{figure}[!htb]\centering
\includegraphics[width=0.95\linewidth]{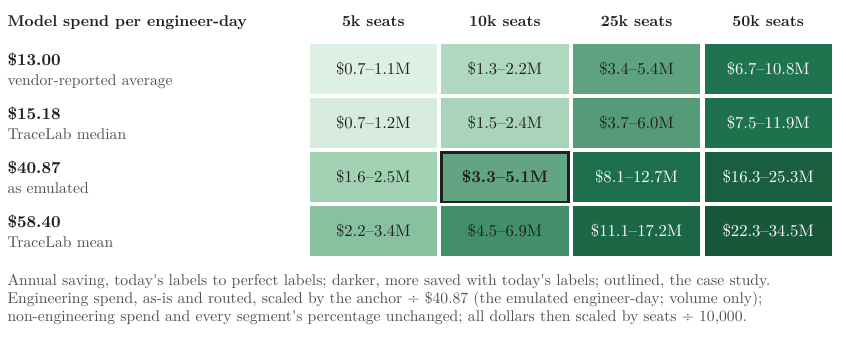}
\caption{Annual saving by model spend per engineer-day and by seats, from today's labels to perfect labels; the outlined cell is the case study. Engineering spend scales with the spend level; other spend and every segment's percentage stay as emulated.}\label{fig:sizing}
\end{figure}

%% file: figures/tab_sensitivities.tex
\begin{tabular}{lrrr}
\toprule
Case (population, same labels) & As-is / month & Routed / month & Saving \\
\midrule
Base case & \$2.03M & \$1.76M & 13.4\% \\
Ledger rebase at task boundaries & \$2.03M & \$1.76M & 13.4\% \\
One-hour cache lifetime (as-is and routed) & \$2.01M & \$1.70M & 15.5\% \\
No Fable 5.1 licence (as-is on Opus 5, router capped at Opus 5) & \$2.07M & \$2.01M & 3.1\% \\
Everyone as-is on Opus 5 (Fable 5.1 licensed for the router) & \$2.07M & \$1.76M & 15.0\% \\
Confidence threshold 0.7 (base 0.8) & \$2.03M & \$1.84M & 9.6\% \\
Confidence threshold 0.9 (base 0.8) & \$2.03M & \$1.75M & 13.7\% \\
Uniform 0.1$\times$ cache reads, writes 1.25$\times$ / 2$\times$ & \$2.37M & \$2.23M & 5.9\% \\
Uniform 0.1$\times$ cache reads, no write premium & \$2.27M & \$2.14M & 5.6\% \\
Agentic intensity $\times$0.5 & \$1.09M & \$1.09M & 0.3\% \\
Agentic intensity $\times$2 & \$4.53M & \$3.45M & 23.8\% \\
Every answer accepted (no correction turns) & \$1.31M & \$1.12M & 13.8\% \\
Pauses as authored (almost none over five minutes) & \$1.82M & \$1.40M & 22.8\% \\
Pauses as in TraceLab (27\% over five minutes) & \$2.17M & \$2.00M & 8.2\% \\
No subagents beyond the authored ones & \$1.96M & \$1.73M & 12.1\% \\
Subagents at the SWE-chat rate (16\% of tool turns) & \$2.16M & \$1.82M & 15.5\% \\
Misroutes cost a single redo & \$2.03M & \$1.74M & 14.2\% \\
Subagents inherit the parent's model (no launch routing) & \$2.03M & \$1.80M & 11.2\% \\
As-is Fable 5.1 share as observed (3\% of engineering sessions) & \$2.06M & \$1.76M & 14.7\% \\
As-is Fable 5.1 for any complex or expert work (45 / 35 / 14\%) & \$1.98M & \$1.76M & 11.1\% \\
Non-engineers pick Fable 5.1 for half their sessions & \$2.08M & \$1.77M & 14.8\% \\
Non-engineers pick Fable 5.1 for all their sessions & \$2.13M & \$1.78M & 16.3\% \\
\bottomrule
\end{tabular}